\documentclass[10pt,twocolumn,letterpaper]{article}

\usepackage{iccv}
\usepackage{times}
\usepackage{epsfig}
\usepackage{graphicx}
\usepackage{amsmath}
\usepackage{amssymb}
\usepackage{booktabs}
\usepackage{bbding} 
\usepackage{bm}
\usepackage{enumerate}
\usepackage{multirow}
\usepackage{multicol}
\usepackage{makecell}
\usepackage{colortbl}
\usepackage{color}
\usepackage{subcaption}
\usepackage{float}
\usepackage{soul}
\usepackage[noend]{algorithm2e}
\usepackage[pagebackref=true,breaklinks=true,letterpaper=true,colorlinks,bookmarks=false]{hyperref}

\iccvfinalcopy 

\def\iccvPaperID{5821} 

\ificcvfinal\fi

\newcommand{\Paragraph}[1]{\vspace{1mm} \noindent \textbf{#1} \hspace{0mm}}
\renewcommand\footnotemark{}

\begin{document}

\title{AccFlow: Backward Accumulation for Long-Range Optical Flow}
\author{
Guangyang Wu$^{1,2*}$\quad
Xiaohong Liu$^{1 \dagger}$\quad 
Kunming Luo$^{5 \ddagger}$\quad
Xi Liu$^{2 \ddagger}$\quad
Qingqing Zheng$^3$\\
Shuaicheng Liu$^2$\quad
Xinyang Jiang$^4$\quad
Guangtao Zhai$^1$\quad
Wenyi Wang$^{2 \dagger}$\\
$^1$Shanghai Jiao Tong University\quad
$^2$University of Electronic Science and Technology of China\\
$^3$Shenzhen Institute of Advanced Technology\quad
$^4$Microsoft Research Asia\\
$^5$Hong Kong University of Science and Technology
\thanks{$^\dagger$ Corresponding authors. $^\ddagger$ Equal contribution. $^*$ Work was partially finished at the University of Electronic Science and Technology of China.}
}

\maketitle
\ificcvfinal\thispagestyle{empty}\fi

\begin{abstract}
   Recent deep learning-based optical flow estimators have exhibited impressive performance in generating local flows between consecutive frames. However, the estimation of long-range flows between distant frames, particularly under complex object deformation and large motion occlusion, remains a challenging task. One promising solution is to accumulate local flows explicitly or implicitly to obtain the desired long-range flow. Nevertheless, the accumulation errors and flow misalignment can hinder the effectiveness of this approach. This paper proposes a novel recurrent framework called AccFlow, which recursively backward accumulates local flows using a deformable module called as AccPlus. In addition, an adaptive blending module is designed along with AccPlus to alleviate the occlusion effect by backward accumulation and rectify the accumulation error. Notably, we demonstrate the superiority of backward accumulation over conventional forward accumulation, which to the best of our knowledge has not been explicitly established before. To train and evaluate the proposed AccFlow, we have constructed a large-scale high-quality dataset named CVO, which provides ground-truth optical flow labels between adjacent and distant frames. Extensive experiments validate the effectiveness of AccFlow in handling long-range optical flow estimation. Codes are available at \url{https://github.com/mulns/AccFlow}.
\end{abstract}

\section{Introduction}
\label{sec:intro}

Optical flow is ideally a dense field of motion vectors that depicts the pixel-wise correspondence of two video frames. Since a variety of downstream applications (\eg, video editing~\cite{bonneel2015blind,huang2022rife,YAN201911}, action recognition~\cite{Simonyan2014}, and object tracking~\cite{behl2017bounding}) significantly benefit from the accuracy of flow estimation, optical flow estimation turns out to be a long-standing fundamental task in computer vision~\cite{vfiformer,stvsr,gdcvfi,rawvsr,vsr,ovssr,mfsr,mrsr}.

Recent advances~\cite{Flownet_flyingchairs,pwc_net,raft2020} resort to deep learning to estimate optical flow and achieve promising accuracy. Although remarkable performance has been achieved in the \textit{local} flow estimation between two adjacent frames, it is non-trivial to estimate the \textit{long-range} flow that records the pixel correspondence between two distant frames.

\begin{figure}
   \includegraphics[width=\linewidth]{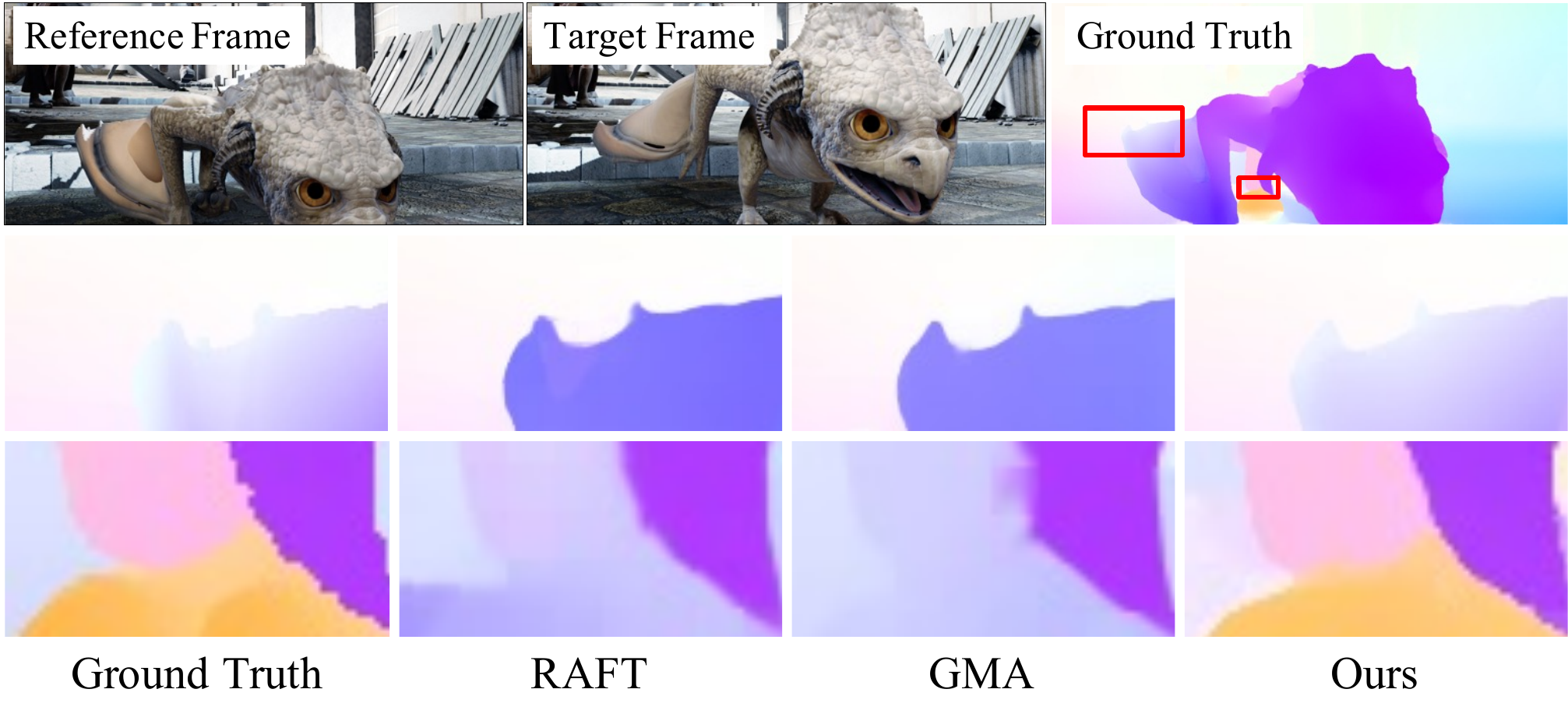}
   \centering
   \caption{
      Comparisons of our method with RAFT~\cite{raft2020} and GMA~\cite{jiang2021learning} on {HS-Sintel dataset}~\cite{janai2017slow}. Zoom-in regions are annotated in red boxes. Our method outperforms other methods especially for occluded area.
   }
  \vspace{-0.5cm}
   \label{fig:teaser}
\end{figure}

The long-range optical flow is a grounded research topic that has plenty of practical applications. For instance, in video completion~\cite{gao2020vc}, long-range optical flow is beneficial to the detail compensation between distant frames;
in video key-point propagation~\cite{harley2022particle}, since the long-range optical flow performs holistic pixel-tracking in nature, it frees the quantity limitation of tracked pixels; in video super-resolution~\cite{luo2022bsrt}, it enables better inter-frame alignment in one sliding window; and in segmentation mask propagation~\cite{wang2019learning}, it provides an explicit approach to propagate masks to distant frames, improving the interpretability compared to the implicit matching. The above examples take a glance at the wide applications of long-range optical flow. More significantly, the success of this task has the potential to break through the performance bottleneck of relevant tasks. 


Surprisingly, even though the long-range optical flow is significant and can benefit many related tasks, few works put effort on this research line. One possible reason is the lack of public datasets that provide ground-truth bidirectional cross-frame optical flows for training and validation. In literature, {the early attempt} to address this long-range optical flow task is Lim~\etal~\cite{LimAG05}, which proposed a method based on the \textit{forward} flow accumulation, in which the flows of adjacent frames are added successively along the motion trajectories. The recent work~\cite{janai2017slow} follows this idea and reasons the occlusion regions from high-frame-rate frames. Apart from these, one can simply estimate the long-range flow by employing methods specified for local flow~\cite{jiang2021learning,raft2020}. As shown in Figure~\ref{fig:teaser}, since the influence of occlusion is {positively related} to the time interval between two frames, the accuracy of flow estimation from these methods would be deteriorated severely or even be unacceptable when the time interval is beyond a threshold. In addition, one can also traverse all pixels in a frame and employ the pixel-tracking methods~\cite{sand2008particle,harley2022particle} to produce the long-range dense flow, which has huge computational overheads and {cannot be used} in applications requiring dense flow. To sum up, a considerate long-range optical flow method should address the following challenging issues: 

\begin{enumerate}[1)]
\item\textbf{Occlusion} As the time interval increases, the flow estimation of two distant frames suffers significant degradation owing to the inter-frame occlusion. Therefore, without a specific design, the common methods that aim at dealing with local flows perform poorly. Janai~\etal~\cite{janai2017slow} formulate it as an energy minimization problem and found it highly non-convex, so they exploit the linearity of small motions and reasons about occlusions from multiple frames. However, this strategy is based on high-frame-rate videos ($\geq 240$ FPS) and not applicable on regular videos.

\item\textbf{Accumulation error} Although flow accumulation is a promising solution to tackle long-range flow estimation, it also brings the accumulation error, resulting in inaccurate estimation in non-occluded regions. Therefore, the effectiveness of accumulation error compensation is critical. Lim~\etal~\cite{LimAG05} and Janai~\etal~\cite{janai2017slow} constrained the photo consistency of warped frames to shrink accumulated error. However, the photo consistency loss is not comprehensive for flow estimation as revealed in~\cite{janai2018unsupervised,Liu2019CVPR}.

\item\textbf{Efficiency} The computational complexity of long-range optical flow should be controlled at an appropriate level to support the downstream tasks in practice. Therefore, the pixel-tracking methods~\cite{sand2008particle,harley2022particle}, which iterative estimate the per-pixel long-range displacement, do not satisfy this requirement.
\end{enumerate}

To address the above issues, we propose a novel framework, named AccFlow, to estimate long-range optical flow by progressively backward accumulating local flows with effective corrections. More specifically, to alleviate the occlusion effect, we propose the \textit{backward accumulation}, a new accumulation strategy distinct from the \textit{forward accumulation} pipeline, and elaborate a corresponding deep module, named AccPlus. More details about the difference between backward and forward accumulation can be found in Section~\ref{subsec:forward} and \ref{subsec:backward}. The AccFlow framework consists of three components: an arbitrary optical flow estimator, the AccPlus module, and an adaptive blending module. The arbitrary optical flow estimator is used to estimate local flows and long-range initial flow. The AccPlus performs the backward accumulation in feature domain. The adaptive blending module rectifies the accumulated error. 
Furthermore, to train and validate our AccFlow, we elaborately build a large-scale synthetic dataset, named CVO (cross-frame video optical flows). Different from other synthetic flow datasets~\cite{butler2012naturalistic,Flownet_flyingchairs}, the CVO includes {\textit{comprehensive}} cross-frame bidirectional flow annotations. The CVO also includes more challenging cases that have large pixel displacement and severe occlusion.

The contributions of this paper can be summarized as follows:

$\bullet$ We propose a novel \textbf{backward accumulation} strategy to alleviate the long-range occlusion.

$\bullet$ We build the CVO, a new large-scale synthetic dataset with comprehensive \textbf{cross-frame} optical flow annotations.

$\bullet$ We propose the \textbf{AccFlow framework} which is simple yet effective to predict the long-range optical flow and achieves the state-of-the-art results on several benchmarks. 

\section{Related Works}
\label{sec:related}

\subsection{Adjacent Frame Optical Flow Estimation}
Optical flow methods can be categorized into two-frame and multi-frame methods according to the number of input frames. For two-frame methods, traditional algorithms~\cite{Thomas2004,Sun2010,EpicFlow_2015} obtain optical flow by minimizing well-designed energy functions based on the brightness constancy assumption. By training a convolutional network on a synthetic dataset, FlowNet~\cite{Flownet_flyingchairs} first established a deep learning approach for optical flow estimation. After that, the performance of optical flow estimation is gradually improved by various works, such as FlowNet2~\cite{ilg2017flownet}, PWC-Net~\cite{pwc_net}, and IRR-PWC~\cite{hur2019iterative}. Recently, RAFT~\cite{raft2020} proposed a new paradigm to estimate optical flow by introducing 4D correlation volume and recurrent network. Following RAFT, graph reasoning~\cite{ag2022learning}, global motion aggregation~\cite{jiang2021learning}, kernel patch attention~\cite{luo2022learning}, and cross-attention transformer~\cite{Sui_2022_CVPR} are further proposed to improve the accuracy and efficiency. 

The purpose of multi-frame optical flow estimation is to estimate the optical flow of adjacent frames by utilizing the temporal information of multiple video frames. Traditional methods achieve multi-frame optical flow estimation by phase-based representations of local image structure~\cite{fleet1990computation,heeger1988optical}, spatial-temporal regularization term~\cite{janai2017slow,weickert2001variational,stoll2013joint}, constant velocity prior~\cite{janai2017slow,volz2011modeling,salgado2007temporal,sun2010layered,wang2008estimating}, constant acceleration assumption~\cite{black1991robust,kennedy2015optical}, and directional prior~\cite{maurer2018directional}. Recently, deep-based multi-frame methods are proposed to fuse flow prediction~\cite{ren2019fusion} or feature~\cite{neoral2018continual,godet2021starflow} from previous frame pair into the current estimation process. 

Although these optical flow methods have achieved remarkable performance, they mainly focus on estimating optical flow of two adjacent frames, leaving the long-range optical flow of non-adjacent frames rarely being explored.
  
\subsection{Non-adjacent Frame Optical Flow Estimation}
Lim~\etal~\cite{lim2001optical} proposed the early work to obtain the cross-frame optical flow, where the Lucas-Kanade method~\cite{lucas1981iterative} is used to produce optical flow at a high frame rate and the accumulation strategy is designed to generate optical flow at a standard frame rate. After that, this accumulation method is improved by accumulation error modeling and correction~\cite{lim2004benefits,LimAG05,janai2017slow}. 
Janai~\etal~\cite{janai2017slow} cast this task as an energy minimization problem, and opt for a data-driven hypothesis generation strategy for optimization.
Recently, Harley~\etal~\cite{harley2022particle} proposed a deep CNN network, PIPs, to estimate cross-frame sparse optical flow from the perspective of per-pixel tracking over the video sequence. Although PIPs has achieved state-of-the-art performance for video pixel tracking, it is difficult to obtain long-range dense optical flow due to the lack of spatial coherence information. 
In this paper, we deeply analyze the drawbacks of existing accumulation strategies and propose a new accumulation framework for obtaining long-range dense optical flow.

\section{Methods}
\label{sec:method}

Let $\mathcal{I} = \{\mathbf{I}_1, \dots, \mathbf{I}_N\}$ denote a video sequence with $N$ image frames $\mathbf{I}_t \in \mathbb{R}^{w\times h\times 3}$ of size $w\times h$ and $3$ color channels. Let $\mathbf{F}_{i, j} \in \mathbb{R}^{w\times h\times 2}$ denote the optical flow field from the reference image $\mathbf{I}_i$ to the target image $\mathbf{I}_j$. Specifically, for each pixel $\mathbf{x} \in \Omega_i=\left\{ 1,\dots,w \right\}\times \left\{ 1,\dots,h \right\}$ in reference image $\mathbf{I}_i$, $\mathbf{F}_{i, j}(\mathbf{x}) \in \mathbb{R}^2$ describes the apparent motion from frame $I_i$ to $I_j$.

Our goal is to estimate the long-range optical flow field $\mathbf{F}_{1, N}$ by accumulating all intermediate local flow fields $\left\{ \mathbf{F}_{1, 2}, \dots, \mathbf{F}_{N-1, N} \right\}$. To achieve this, Lim~\etal~\cite{LimAG05} and Janai~\etal~\cite{janai2017slow} formulate it as a dense pixel tracking task and obtain the long-range flow by tracking through pixel trajectories. In this paper, we refer to these approaches as the \textit{forward accumulation}. In Section~\ref{subsec:forward}, we revisit the forward accumulation process and provide a formalization of it. The essential problem inherent in this process is analyzed, and a solution referred to as \textit{backward accumulation} is proposed in Section~\ref{subsec:backward}. Subsequently, we introduce in Section~\ref{subsec:accflow} the proposed AccFlow framework that accomplishes the aforementioned backward accumulation to mitigate the occlusion effect and rectify the accumulated error. Additionally, we introduce the proposed CVO dataset which provides synthesized video with ground-truth long-range optical flow between distant frames in Section~\ref{subsec:dataset}.

\subsection{Revisiting the Forward Accumulation}
\label{subsec:forward}

Generally, the accumulation process is a recursive procedure to fuse all intermediate local flows together. For brevity, we define the fusion of two adjacent optical flows $\mathbf{F}_{i, k}$ and $\mathbf{F}_{k, j}$ as $\oplus$, and we present the fused flow $\mathbf{F}_{i,j}$ as:
\begin{equation}
   \label{eq:fuse}
   \mathbf{F}_{i, j} = \mathbf{F}_{i, k} \oplus \mathbf{F}_{k, j}
\end{equation}
where $i,k,j\in [1,N]$ denote three time stamps satisfying $i<k<j$.
Since the adjacent flows $\mathbf{F}_{i, k}$ and $\mathbf{F}_{k, j}$ start at different frames (\ie, frame $\mathbf{I}_i$ and $\mathbf{I}_k$), in order to obtain the target flow $\mathbf{F}_{i,j}$ which starts at frame $\mathbf{I}_i$, we need to warp the start point of each motion vector in $\mathbf{F}_{k, j}$ to align them with $\mathbf{F}_{i, k}$, and then add the two flows pixel-wise. Let $\widetilde{\mathbf{F}}_{k, j}^i$ denote the warped $\mathbf{F}_{k, j}$ starting at frame $\mathbf{I}_i$, we have:
\begin{equation}
   \label{eq:warp}
   \widetilde{\mathbf{F}}_{k, j}^i(\mathbf{x}) = \mathbf{F}_{k, j}(\mathbf{x}+\mathbf{F}_{i, k}(\mathbf{x}))
\end{equation}
for each pixel $\mathbf{x}$ in reference image $\mathbf{I}_i$. Then we obtain the target flow $\mathbf{F}_{i,j}$ by:
\begin{equation}
   \label{eq:warp-fuse}
   \mathbf{F}_{i, j}(\mathbf{x}) = \mathbf{F}_{i, k}(\mathbf{x}) + \widetilde{\mathbf{F}}_{k, j}^i(\mathbf{x}).
\end{equation}

However, as Janai~\etal~\cite{janai2017slow} revealed, the reference pixel $\mathbf{x}\in \Omega_i$ can be forward occluded in frame $\mathbf{I}_k$, which leads to wrong warping results in Equation~(\ref{eq:fuse})-(\ref{eq:warp-fuse}). Therefore, researchers usually speculate on the occlusion mask and solve the occluded regions by estimation. For brevity, we define the binary occlusion mask $\mathbf{O}_{i,k}$, where $\mathbf{O}_{i,k}(\mathbf{x}) \in \left\{ 0,1 \right\}$ specifies whether pixel $\mathbf{x}\in \Omega_i$ is forward occluded from frame $\mathbf{I}_i$ to $\mathbf{I}_k$. Equation~(\ref{eq:fuse})-(\ref{eq:warp-fuse}) valid only when pixel $\mathbf{x}\in \Omega_i$ is not occluded in frame $I_k$ (\ie, $\mathbf{O}_{i,k}(\mathbf{x})=0$). As for occluded pixels  (\ie, $\mathbf{O}_{i,k}(\mathbf{x})=1$), its optical flow has to be estimated by some carefully designed  occlusion solvers. For easy notation, function $\mathit{solveOcc}$ denotes occlusion solvers in general , and $\mathbf{P}_{i,j}\in \mathbb{R}^{w\times h\times 2}$ denote the estimated flows in occluded region, where
\begin{equation}
   \label{eq:solveocc}
   \mathbf{P}_{i,j} = \mathit{solveOcc}(\mathbf{F}_{i, k}, \mathbf{F}_{k, j}, \mathbf{O}_{i,k}).
\end{equation}
Therefore, Equation~(\ref{eq:warp-fuse}) can be re-formulated as:
\begin{equation}
   \label{eq:plus}
   \mathbf{F}_{i, j}(\mathbf{x}) =
      \begin{cases}
         \mathbf{F}_{i, k}(\mathbf{x}) + \widetilde{\mathbf{F}}_{k, j}^i(\mathbf{x})  & \text{if } \mathbf{O}_{i, k}(\mathbf{x}) = 0, \\
         \mathbf{P}_{i,j}(\mathbf{x})   & \text{if } \mathbf{O}_{i, k}(\mathbf{x}) = 1.
      \end{cases}
\end{equation}

The forward accumulation process recursively performs the above operations. Specifically, with the time index $t$ increases from $2$ to $N-1$, we recursively produce $\mathbf{F}_{1, t+1}$ by fusing the pre-obtained flow $\mathbf{F}_{1, t}$ and the local flow $\mathbf{F}_{t, t+1}$ as follows:
\begin{equation}
   \mathbf{F}_{1, t+1} = \mathbf{F}_{1, t} \oplus \mathbf{F}_{t, t+1},
\end{equation}
where for each pixel $\mathbf{x}\in \Omega_1$ in reference image $\mathbf{I}_1$, we have
\begin{equation}
   \label{eq:forward-plus}
   \mathbf{F}_{1, t+1}(\mathbf{x}) =
      \begin{cases}
         \mathbf{F}_{1, t}(\mathbf{x}) + \widetilde{\mathbf{F}}_{t, t+1}^1(\mathbf{x})  & \text{if } \mathbf{O}_{1, t}(\mathbf{x}) = 0, \\
         \mathbf{P}_{1,t+1}(\mathbf{x})   & \text{if } \mathbf{O}_{1, t}(\mathbf{x}) = 1,
      \end{cases}
\end{equation}
where the occlusion mask $\mathbf{O}_{1,t}$ is usually estimated as well. We denote the occlusion reasoning methods as $\mathit{getOcc}$ in general:
\begin{equation}
   \mathbf{O}_{1, t} = \mathit{getOcc}(\mathbf{F}_{1,t}, \mathbf{F}_{t,t+1}).
\end{equation}
For clarity, we present the pseudocode of the forward accumulation process in Algorithm~\ref{alg:forward}.

\RestyleAlgo{ruled}
\SetStartEndCondition{ }{ }{ }
\SetKwFor{For}{for}{\string:}{}%
\SetKwIF{If}{ElseIf}{Else}{if}{:}{elif}{else:}{}%
\SetKwFor{While}{while}{}{}%
\SetKwRepeat{Repeat}{repeat}{until}%
\AlgoDisplayBlockMarkers
\SetAlgoNoLine%
\AlgoDontDisplayBlockMarkers\SetAlgoNoEnd\SetAlgoNoLine%

\begin{algorithm}[tbp]
   \caption{The Forward Accumulation}\label{alg:forward}
   \KwIn{$\left\{ \mathbf{F}_{t,t+1} \mid t \in [1,N-1] \right\}$}
   \KwOut{$\mathbf{F}_{1,N}$}
   \For{$t\gets 2$ \KwTo $N-1$}{
      $\mathbf{O}_{1,t}\gets \mathit{getOcc}(\mathbf{F}_{1,t},\mathbf{F}_{t,t+1})$\\
      $\mathbf{P}_{1,t+1} \gets \mathit{solveOcc}(\mathbf{F}_{1, t}, \mathbf{F}_{t, t+1}, \mathbf{O}_{1, t})$\\
      \For{$\mathbf{x} \in \Omega_1$}{
         $\widetilde{\mathbf{F}}_{t,t+1}^1(\mathbf{x}) \gets \mathbf{F}_{t, t+1}(\mathbf{x}+\mathbf{F}_{1, t}(\mathbf{x}))$\\
         \If{$\mathbf{O}_{1, t}(\mathbf{x}) = 0$}{
            $\mathbf{F}_{1, t+1}(\mathbf{x}) \gets \mathbf{F}_{1, t}(\mathbf{x}) + \widetilde{\mathbf{F}}_{t, t+1}^1(\mathbf{x})$
         }{\ElseIf{$\mathbf{O}_{1, t}(\mathbf{x}) = 1$}{
               $\mathbf{F}_{1, t+1}(\mathbf{x})\gets \mathbf{P}_{1,t+1}(\mathbf{x}) $
            }
         }
      }
   }
\end{algorithm}

\subsection{Backward Accumulation}
\label{subsec:backward}

Previous research~\cite{janai2017slow} has shown that the forward accumulation can generate high quality motion hypotheses for visible regions, but the occluded regions limit its performance. In this subsection, we first analyze the occlusion area in the forward accumulation process, then propose a new solution to alleviate the occlusion effect.

Let $\Delta=|k-i| \ge 1$ denote the time interval, we define the proportion of occluded area of $\mathbf{O}_{i,k}$ as:
\begin{equation}
   \label{eq:alpha-define}
   \alpha_{\Delta}^i = \frac{\sum_{\mathbf{x}\in \Omega_i} \mathbf{O}_{i,k}(\mathbf{x})}{h\times w},
\end{equation}
where $\alpha^i_{\Delta} \in [0,1]$. 
We begin by analyzing the case of a one-dimensional object moving with constant velocity, assuming that the object is of length $\delta  w$ pixels, the canvas length is $M\gg \delta w$, the velocity of the object is $v$ pixels per frame, and the background is fixed. From time $t=1$ to $t=k$, the proportion of forward occluded area is calculated as:
\begin{equation}
   \label{eq:alpha}
   \alpha_{|k-1|}^1 = \frac{ \min\{v\times |k-1|, \delta w \} }{M},
\end{equation}
which is positively correlated with the time interval $|k-1|$. Similar conclusions can be extended to two-dimensional cases. Thus, the inequality 
\begin{equation}
   \label{eq:inequality}
   \alpha_{\Delta + 1}^i \geq \alpha_{\Delta }^i,
\end{equation}
holds for linear motion. 

\begin{algorithm}[tbp]
   \caption{The Backward Accumulation}\label{alg:backward}
   \KwIn{$\left\{ \mathbf{F}_{t,t+1} \mid t \in [1,N-1] \right\}$}
   \KwOut{$\mathbf{F}_{1,N}$}
   \For{$t\gets N-1$ \KwTo $2$}{
      $\mathbf{O}_{t-1,t} \gets \mathit{getOcc}(\mathbf{F}_{t-1,t},\mathbf{F}_{t,N})$\\
      $\mathbf{P}_{t-1,N} \gets \mathit{solveOcc}(\mathbf{F}_{t-1, t}, \mathbf{F}_{t, N}, \mathbf{O}_{t-1, t})$\\
      \For{$\mathbf{x} \in \Omega_{t-1}$}{
         $\widetilde{\mathbf{F}}^{t-1}_{t, N}(\mathbf{x}) \gets \mathbf{F}_{t, N}(\mathbf{x}+\mathbf{F}_{t-1, t}(\mathbf{x}))$\\
         \If{$\mathbf{O}_{t-1, t}(\mathbf{x}) = 0$}{
            $\mathbf{F}_{t-1, N}(\mathbf{x}) \gets \mathbf{F}_{t-1, t}(\mathbf{x}) + \widetilde{\mathbf{F}}^{t-1}_{t, N}(\mathbf{x})$
         }{\ElseIf{$\mathbf{O}_{1, t}(\mathbf{x}) = 1$}{
               $\mathbf{F}_{t-1, N}(\mathbf{x})\gets  \mathbf{P}_{t-1, N}(\mathbf{x}) $\
            }
         }
      }
   }
\end{algorithm}

\begin{figure}[tbp]
   \includegraphics[width=\linewidth]{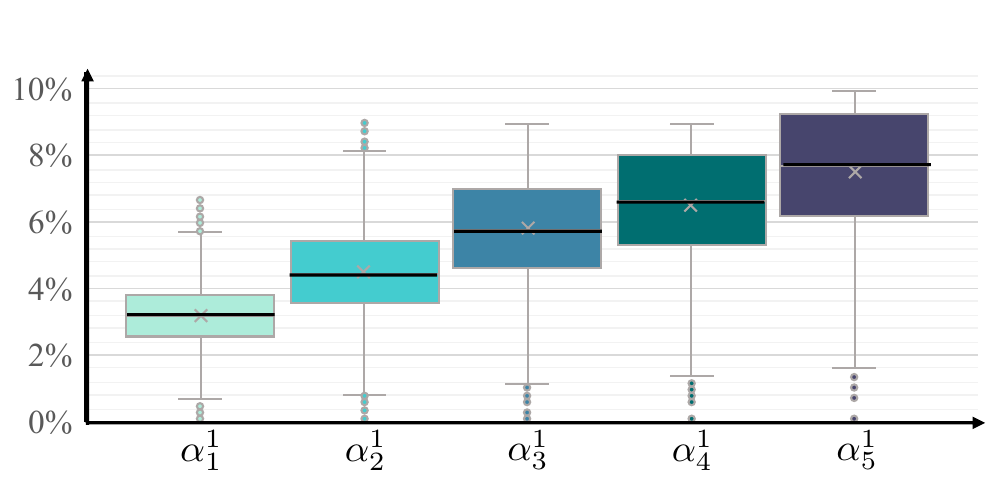}
   \centering
   \caption{Box-plot of the occlusion proportion $\alpha_\Delta^1$ over 5000 samples, the occlusion proportion (Y-axis) increases with the time interval $\Delta$ (X-axis) increases.}
   \label{fig:box}
\end{figure}

While the assumption of linear motion may not always hold in practical scenarios, our experiments show that Equation~(\ref{eq:inequality}) remains valid when a significant number of samples are tested. The statistical results over 5000 samples are provided in terms of box-plot in Figure~\ref{fig:box}, which demonstrates that the $\alpha_\Delta^i$ is positively correlated with $\Delta$ as Equation~(\ref{eq:alpha}) indicates. This conclusion is important for the following analysis.

Algorithm~\ref{alg:forward} shows that the occlusion proportion $\alpha_{t-1}^1$ of $\mathbf{O}_{1,t}$ increases progressively with $t$ increases, which significantly burdens the  occlusion solver. Although existing techniques~\cite{jiang2021learning,raft2020} can powerfully solve occlusion with deep neural networks (DNN), the constant increment of the occlusion proportion is still a challenge that might consume substantial computational resources.

\begin{figure}[tbp]
   \includegraphics[width=\linewidth]{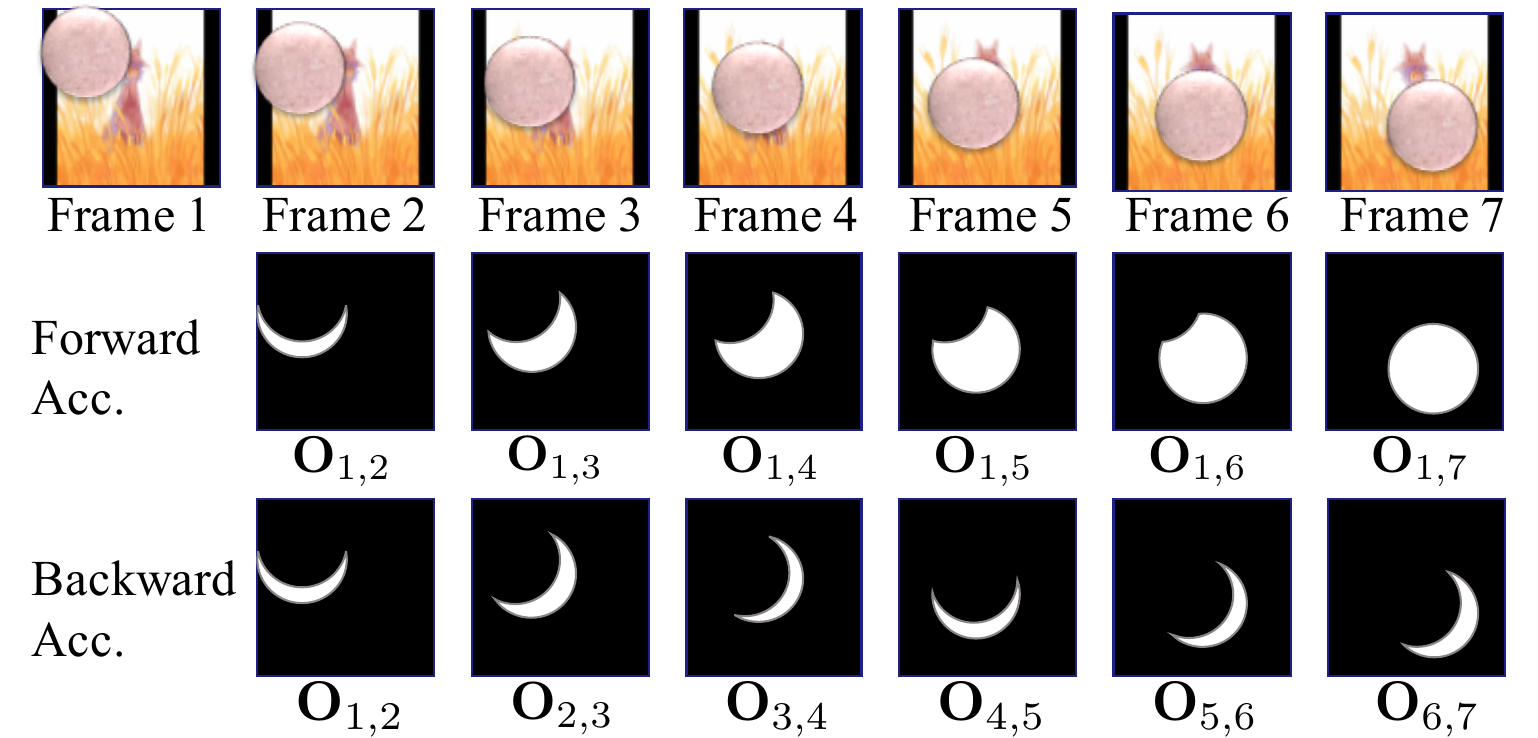}
   \centering
   \caption{
      Visualization of occlusion masks during accumulation. White regions denote occluded area. 
   }
   \vspace{-0.5cm}
   \label{fig:occ}
\end{figure}

To address the above critical issue, we propose a simple solution, named the \textit{backward accumulation}, where we reverse the accumulation order without extra computational complexity involved. As analyzed in Equation~(\ref{eq:warp-fuse})-(\ref{eq:solveocc}), the alignment operation introduces errors in the forward occluded regions, and as revealed in Equation~(\ref{eq:inequality}), they are proportionally correlated with the time interval. In each step of accumulation process, we can simplify the problem as the alignment of two optical flows, one of which has a larger magnitude (pre-obtained from the last step) and another one has a smaller magnitude (the local flow). The forward accumulation chooses to align two flows along the larger one, which essentially leads to a larger occlusion area. Therefore, we propose to align the two flows along the smaller one. Specifically, with time variable $t$ decreases from $N-1$ to $2$, we recursively produce the long-range flow $\mathbf{F}_{t-1, N}$ by fusing the pre-obtained flow $\mathbf{F}_{t, N}$ and the local flow $\mathbf{F}_{t-1, t}$ as follows:
\begin{equation}
   \mathbf{F}_{t-1, N} = \mathbf{F}_{t-1, t} \oplus \mathbf{F}_{t, N},
\end{equation}
where for each pixel $\mathbf{x}\in \Omega_{t-1}$ in reference image $\mathbf{I}_{t-1}$, we have
\begin{equation}
   \label{eq:backward-plus}
   \mathbf{F}_{t-1, N}(\mathbf{x})=
      \begin{cases}
         \mathbf{F}_{t-1, t}(\mathbf{x}) + \widetilde{\mathbf{F}}_{t, N}^{t-1}(\mathbf{x})  & \text{if } \mathbf{O}_{t-1, t}(\mathbf{x}) = 0, \\
         \mathbf{P}_{t-1,N}(\mathbf{x})   & \text{if } \mathbf{O}_{t-1, t}(\mathbf{x}) = 1,
      \end{cases}
\end{equation}
and the occlusion mask is obtained by:
\begin{equation}
   \mathbf{O}_{t-1, t} = \mathit{getOcc}(\mathbf{F}_{t-1, t},\mathbf{F}_{t, N}).
\end{equation}
By doing this, we form the backward accumulation process presented in Algorithm~\ref{alg:backward}.

As evident from the recursive process, the occluded regions are pixels with $\mathbf{O}_{t-1,t}(\mathbf{x})=1, \mathbf{x}\in \Omega_{t-1}$, at each step. The occlusion proportion defined in Equation~(\ref{eq:alpha-define}) is $\alpha_1^{t-1}$ here.
During the backward accumulation, although the reference image undergoes changes, the occluded region remains at a minimum level, particularly when compared to the forward accumulation method where the occluded region progressively increases.  We visualize this observation in Figure~\ref{fig:occ}. The reduced occluded area enables the occlusion solver to handle the occlusion more efficiently.

\subsection{AccFlow Framework}
\label{subsec:accflow}

\begin{figure}[t]
   \centering
   \begin{subfigure}[t]{\linewidth}
     \centering
     \includegraphics[width=\linewidth]{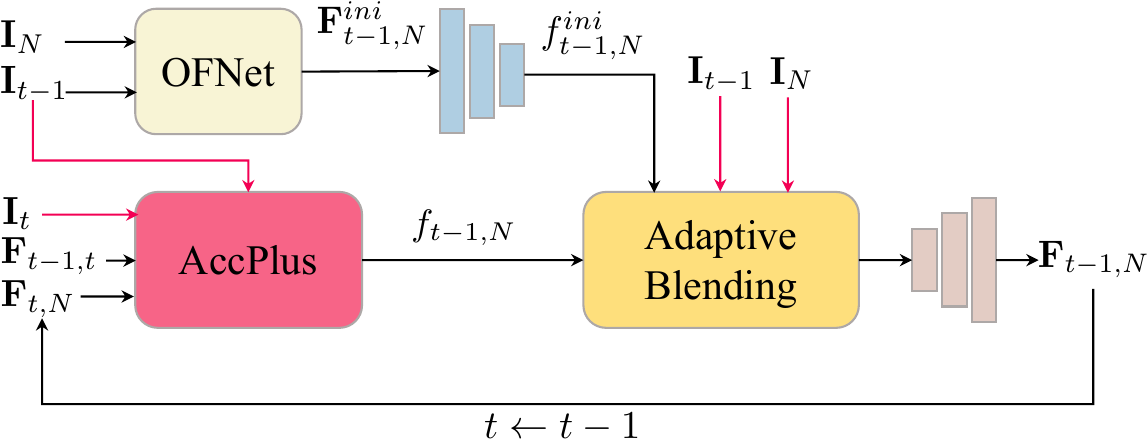}
     \caption{AccFlow Framework.} 
     \label{fig:accflow}
   \end{subfigure}
   \begin{subfigure}[t]{\linewidth}
     \centering
     \includegraphics[width=\linewidth]{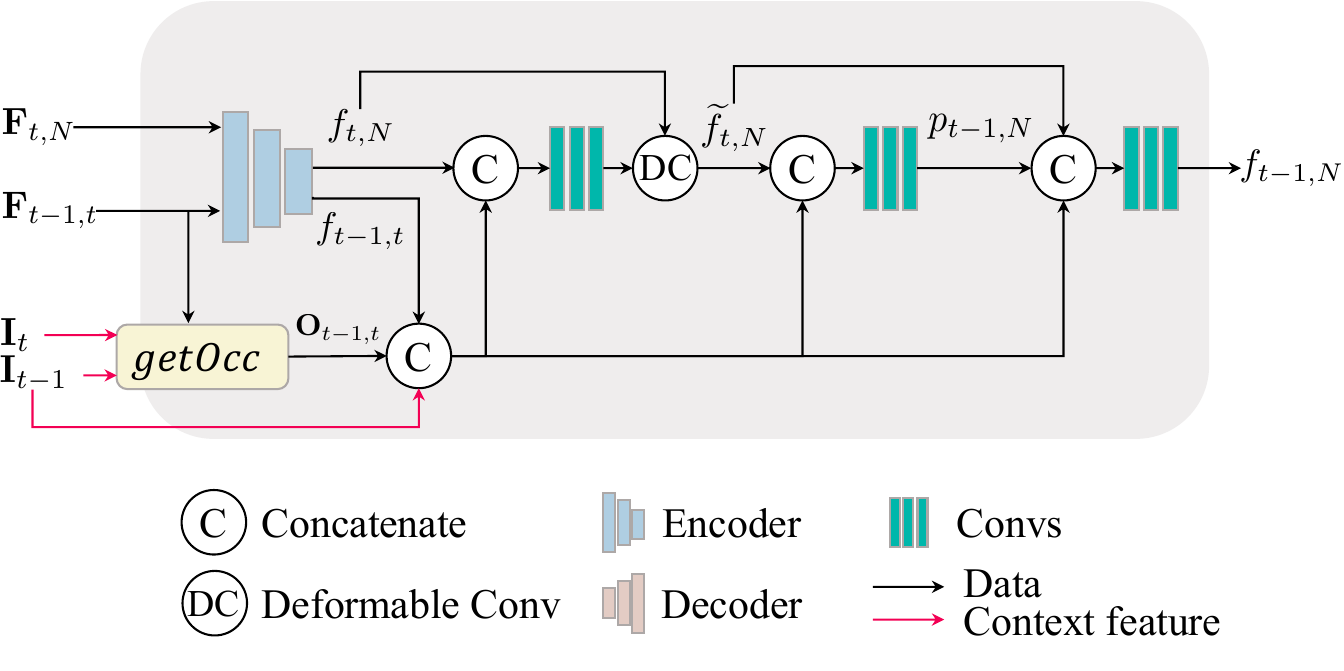}
     \caption{AccPlus Module.}
     \label{fig:accplus}
   \end{subfigure}
   \centering
   \caption{Illustration of the network structure. (a) The AccFlow framework. Time $t$ decreases from $N-1$ to $2$ to obtain long-range flow $\mathbf{F}_{1, N}$. OFNet is an arbitrary flow estimator. (b) The AccPlus module, an efficient module that implements the backward accumulation in feature domain. The \textcolor{red}{red} arrows signify the encoding of images into context features by a context encoder, which adheres to the structure outlined in~\cite{raft2020}.}
   \label{fig:foward&backward}
   \vspace{-0.5cm}
\end{figure}

In this section, we present AccFlow, a deep framework that employs the backward accumulation to estimate accurate long-range optical flow. The framework consists of three components, an arbitrary optical flow estimator OFNet (\eg, RAFT, GMA, \etc), the AccPlus module, and the adaptive fusion module. Initially, local flows $\{\mathbf{F}_{t,t+1} \mid t \in [1,N-1] \}$ are obtained from the pretrained OFNet as inputs of AccFlow. The AccFlow recursively produces the long-range flow $\mathbf{F}_{t-1, N}$ with time $t$ decreases from $N-1$ to $2$ and the recurrent structure is shown in Figure~\ref{fig:accflow}.

\Paragraph{The AccPlus Module.} 
Following the Algorithm~\ref{alg:backward}, we implement the backward accumulation in the AccPlus module to perform flow fusion in feature domain as shown in Figure~\ref{fig:accplus}. At each stage, given the local flow $\mathbf{F}_{t-1,t}$ and pre-obtained flow $\mathbf{F}_{t, N}$, we encode them into motion features $f_{t-1,t}$ and $f_{t, N}$ with a motion encoder. The motion encoder spatially downscales features by $1/4$ times. The occlusion mask $\mathbf{O}_{t-1, t}$ is determined by  $\mathit{getOcc}$ which is a simple  warping operation in this paper. More details about the encoder and $\mathit{getOcc}$ are provided in appendix. Afterwards, we warp the motion features $f_{t, N}$ to align them with $f_{t-1, t}$ by deformable convolution and produce $\widetilde{f}_{t, N}$. 
In the AccPlus, we implement $\mathit{solveOcc}$ in Algorithm~\ref{alg:backward} by a set of convolutional layers.
Specifically, we concatenate $\widetilde{f}_{t, N}$ and ${f}_{t-1, t}$ along the channel dimensional, where ${f}_{t-1, t}$ provides the spatial coherence information for handling occlusion. The concatenated feature is then processed by multiple convolutional layers.
The resulting output features, denoted as $p_{t-1, N}$, are then merged with $\widetilde{f}_{t, N}$ and ${f}_{t-1, t}$ to produce the final target motion feature ${f}_{t-1, N}$.

\Paragraph{The adaptive blending module.} Directly decoding the output features ${f}_{t-1, N}$ of AccPlus and passing them to next stage may result in the accumulation error. To mitigate this issue, an adaptive blending module is added to suppress the accumulation error by using the directly estimated long-range flow as prior information. Specifically, we first establish an initial long-range optical flow $\mathbf{F}_{t-1, N}^{\mathit{ini}}$ with the pretrained OFNet, and then encode it into a motion feature ${f}_{t-1, N}^{\mathit{ini}}$ with the motion encoder (share parameters with the one in AccPlus). Subsequently, the adaptive blending module takes the two motion features (\ie, ${f}_{t-1, N}^{\mathit{ini}}$ and ${f}_{t-1, N}$) and corresponding video frames as inputs to calculate an adaptive confidence mask. The confidence mask is then used to fuse them with attention mechanism, and the output motion features are decoded into the optical flow $\mathbf{F}_{t-1, N}$ with a motion decoder. Details of the motion decoder are provided in appendix.

\subsection{CVO Dataset}
\label{subsec:dataset}

Existing optical flow datasets only provide the local optical flow annotations. In order to provide the ground-truth (GT) long-range optical flows, we construct a cross-frame video optical flow dataset (CVO), consisting of 12K synthetic video sequences and GT optical flow labels across different frame intervals. This dataset is essential for the research on long-range optical flow estimation and other related tasks.

\Paragraph{Dataset Collection}We generate the CVO dataset using Kubric~\cite{greff2022kubric}, which is a data generation pipeline for creating semi-realistic synthetic multi-object videos. We first simulate the movement of multiple objects, and then render frames along with optical flow annotations. For each video sequence, we render 7 frames of size $512\times 512$ at 60 FPS (frame per second) in conjunction with the bidirectional optical flow of adjacent frames. In addition, we provide cross-frame bidirectional optical flows across different frame intervals. All the cross-frame flows take the first frame as reference. We further render the RGB video frames with and without random motion blur, which is denoted as \textit{Clean} and \textit{Final} sets. We partition all video sequences into two subsets, 11K sequences and 500 sequences, which serve as the training and validation splits, respectively.

\begin{figure}[tbp]
   \centering
   \includegraphics[width=1\linewidth]{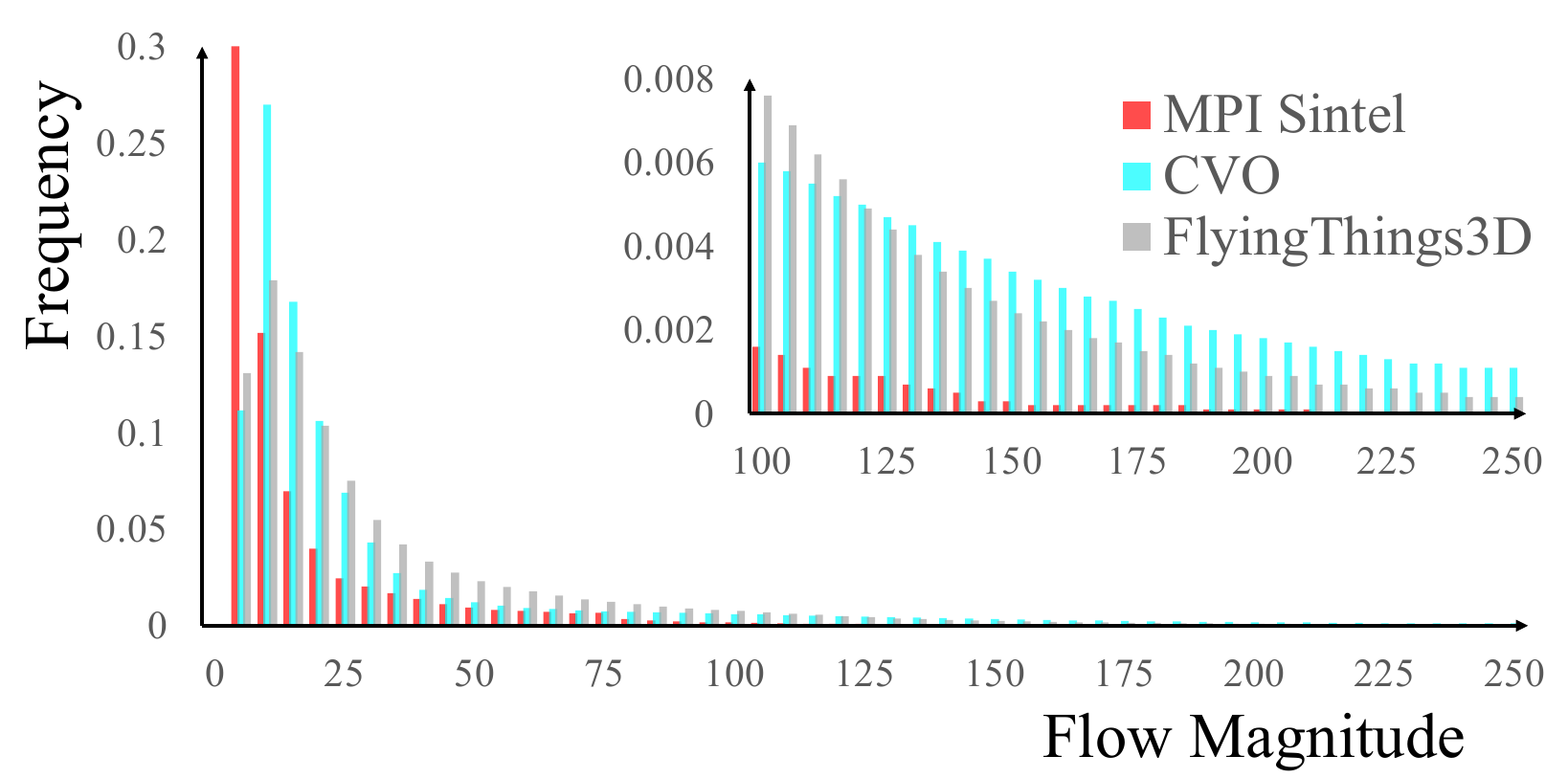}
   \vspace{-0.6cm}
   \caption{The histogram comparisons of the flow magnitude between the training set of CVO and public datasets, such as MPI Sintel~\cite{butler2012naturalistic} and FlyingThings3D~\cite{mayer2016large}. }
   \label{fig:cvo}
\end{figure}

\Paragraph{Comparisons with Existing Datasets}The CVO dataset contains richer annotations compared with existing optical flow datasets~\cite{butler2012naturalistic,Flownet_flyingchairs} since  it provides cross-frame bidirectional optical flow annotations. Moreover, the CVO contains more challenging samples with large motion and complex occlusion. We compare the flow magnitudes among different datasets by plotting the statistical histograms in Figure~\ref{fig:cvo}. Even though the FlyingThings3D~\cite{Flownet_flyingchairs} has similar flow magnitude distribution compared to CVO, the CVO contains more extreme large motions (flow magnitude $\geq 125$ pixels). According to our experiments, the proposed CVO is sufficient to support researches on long-range optical flow estimation and other related tasks.


\section{Experiments}
\label{sec:experiments}


\subsection{Validation Benchmarks}

\Paragraph{CVO:}We adopt the CVO testing set, which consists of \textit{Clean} and \textit{Final} splits, as one of our validation benchmarks. Each split contains $500$ sequences for evaluation. In each sequence, there are $7$ frames of size $512\times 512$, and the default GT optical flow ${\mathbf{F}^{gt}_{1,7}}$. If experiments on other frame intervals are desired, we provide the corresponding GT flow $\mathbf{F}^{gt}_{1, i}, i\in [2,6]$ (denoted as  CVO-$i$).

\Paragraph{HS-Sintel:}MPI Sintel~\cite{butler2012naturalistic} is a commonly used optical flow benchmark generated from the realistic animated film. However, it only provides GT flows at 24 FPS. Therefore, we use the High-Speed Sintel videos~\cite{janai2017slow}, namely HS-Sintel, as an alternative. Specifically, Janai~\etal~\cite{janai2017slow} selected a subset of 19 sequences from the MPI Sintel training set (clean pass) and re-rendered them 24 FPS to 1008 FPS with 4$\times$ resolution. Unfortunately, the GT flows at other frame rates of HS-Sintel are not publicly available. Therefore, we use the GT flows at 24 FPS of MPI Sintel as labels to evaluate the estimates from video sequences at 1008 FPS of HS-Sintel. 
\begin{table*}[t]
	\centering
	\resizebox*{0.86\linewidth}{!}{
		\begin{tabular}
			{
				>{\arraybackslash}p{2.8cm}| 
				>{\centering\arraybackslash}p{1.2cm} 
				>{\centering\arraybackslash}p{1.2cm} 
				>{\centering\arraybackslash}p{1.2cm}| 
        >{\centering\arraybackslash}p{1.2cm} 
				>{\centering\arraybackslash}p{1.2cm} 
				>{\centering\arraybackslash}p{1.2cm}| 
        >{\centering\arraybackslash}p{1.2cm} 
				>{\centering\arraybackslash}p{1.2cm} 
				>{\centering\arraybackslash}p{1.2cm}| 
        >{\centering\arraybackslash}p{1.2cm} 
			}
			\hline
			\multirow{2}{*}{Method} &  \multicolumn{3}{c|}{HS-Sintel} & \multicolumn{3}{c|}{CVO (\textit{Clean})} & \multicolumn{3}{c|}{CVO (\textit{Final})} & Inference   
      \\
			& ALL & NOC & OCC & ALL & NOC & OCC & ALL & NOC & OCC & time (s)   
      \\
      
			\hline
      RAFT &
      2.141 &1.124 &7.169&
      5.687 & 2.798 & 13.233 &
      6.653 & 3.812 & 13.891 &
      0.129 
      \\
      RAFT-\textit{Lim} & 3.868 & 1.845 & 12.63 & 11.96 & 6.573 & 31.10 & 12.34 & 6.938 & 31.45 & 0.956 
      \\
      RAFT-\textit{w} & 1.921  & {1.004} & 6.623 &5.259  &2.274 &12.59  &5.508  &2.493 &12.90 & 0.525 
      \\
      Acc+RAFT (ours) & 
      1.709 & 1.163 & 5.639 &
      3.170 & 1.623 & 8.113 &
      3.283 & 1.714 & 8.261 &
      0.813
      \\

      \hline
      GMA &
      2.291 &1.330 &7.139&
      5.757 & 2.775 & 13.58 &
      6.265 & 3.530 & 13.71 &
      0.234 
      \\
      GMA-\textit{Lim} & 3.871 & 1.764 & 12.79 & 12.22 & 6.708 & 31.40 & 12.42 & 7.038 & 31.61 & 2.159 
      \\
      GMA-\textit{w} & 1.924  & 1.043 & 6.458 &5.136  &2.137 &12.49  &5.515 &2.502 &12.81 & 1.167 
      \\
      Acc+GMA (ours) & 
      1.568 & 1.091 & 5.003 &
      3.583 & 1.807 & 8.868 &
      3.752 & 1.979 & 9.030 &
      1.499
      \\

      \hline
      RAFT$^*$& 
      2.567 &1.426 &7.717&
      4.445 & 1.948 & 11.73 &
      4.537 & 2.003 & 11.70 &
      0.129
      \\
      RAFT$^*$-\textit{Lim} & 3.657 & 1.611 & 12.36 & 23.34 & 6.543 & 32.90 & 13.02 & 7.033 & 33.82 & 0.956 
      \\
      RAFT$^*$-\textit{w} & 2.139 & 1.059 & 6.963 & 3.738  & \textcolor{red}{1.052} &10.41  &3.808 & \textcolor{blue}{1.162} &10.14 & 0.525 
      \\
      Acc+RAFT$^*$ (ours) & 
      \textcolor{red}{1.383} & \textcolor{blue}{0.930} & \textcolor{red}{4.546}  &
      \textcolor{red}{2.634} & 1.155 & \textcolor{red}{7.302} &
      \textcolor{red}{2.707} & 1.249 & \textcolor{red}{7.295} &
      0.813
      \\

      \hline
      GMA$^*$ & 
      2.520 &1.469 &7.600&
      4.638 & 2.342 & 11.33 &
      4.633 & 2.114 & 11.36 &
      0.234 
      \\
      GMA$^*$-\textit{Lim} & 3.306 & 1.381 & 11.70 & 11.39 & 5.833 & 31.28 & 11.68 & 6.130 & 31.35 & 2.159 
      \\
      GMA$^*$-\textit{w} &  1.888 & 0.946 & 6.516 & {3.832} & \textcolor{blue}{1.082} & 10.38  & 3.807  & \textcolor{red}{1.159} & 10.10 & 1.167 
      \\
      Acc+GMA$^*$ (ours) &
      \textcolor{blue}{1.434} & 0.950 & \textcolor{blue}{4.770} &
      \textcolor{blue}{2.732} & 1.181 & \textcolor{blue}{7.438} &
      \textcolor{blue}{2.808} & 1.261 & \textcolor{blue}{7.495} &
      1.499
      \\

      \hline
      SlowFlow & 2.58$^\dagger$  & \textcolor{red}{0.87}$^\dagger$  & 9.45$^\dagger$  & -  &- &- &- &- &- & $\geq$ 500 
      \\
      PIPs & -  & -  & -  & 8.568  & 6.351 & 21.55 & 8.954 & 6.718 & 22.06 & $\geq$ 500
      \\
      GMFlow & 2.055  & 1.024  & 7.132  & 5.801  & 2.680 & 13.521 & 6.506 & 3.402 & 14.21 & 0.341
      \\
			\hline
  \end{tabular}}
  \caption{
    Comparisons of AccPlus framework with other methods on two benchmarks in terms of EPE~$\downarrow$ on all regions (ALL) and occluded regions (OCC). The best and the second-best results are marked in \textcolor{red}{red} and \textcolor{blue}{blue}, respectively. `-\textit{Lim}' denotes the flow accumulation method in~\cite{LimAG05}. `-\textit{w}' denotes the warm-start method (details in Section~\ref{subsec:alternatives}). For the SlowFlow~\cite{janai2017slow}, we refer to data in their paper (denoted with $^\dagger$). We report the inference time of 7 frames of size $512\times 512$ per sample on an NVIDIA GTX3090 GPU.
}
\vspace{-0.38cm}
\label{tab:main}
\end{table*}
\subsection{Implementation Details}

\Paragraph{Loss function:}During the recurrent process to obtain the target flow $\mathbf{F}_{1, N}$, the AccFlow also produces intermediate flows $\mathbf{F}_{t, N}, t\in [1,N-2]$. Therefore, we train the network by supervising all the flow outputs with L1 loss:
\begin{equation}
   \mathcal{L} = \frac{1}{N-2}\sum_{i=1}^{N-2}{\Vert\mathbf{F}_{i, N} - \mathbf{F}^{gt}_{i, N}\Vert}_1.
\end{equation}

\Paragraph{Training details:}We train the AccFlow with the mixture of `clean' and `final' pass of CVO training set. 
We augment the training data by randomly cropping the input frames into patches of size $256\times 256$.
Other training hyperparameters (\textit{e.g.}, learning rate and batch size) follow the default settings from~\cite{raft2020}. 
By replacing the OFNet with different existing optical flow estimators, we train four models for comparison. Specifically, we embed the officially pretrained RAFT~\cite{raft2020} and GMA~\cite{jiang2021learning} in AccFlow framework, respectively. On the one hand, we fix the parameter of OFNet and train other parameters from scratch, and produce Acc+RAFT and Acc+GMA, respectively. On the other hand, we fine-tune the parameter of OFNet and produce Acc+RAFT$^*$ and Acc+GMA$^*$, respectively.

\subsection{Alternative Approaches}
\label{subsec:alternatives}
Previously, several works~\cite{LimAG05,janai2017slow} have been focused on optical flow accumulation. Therefore, for more comprehensive comparisons, we consider some other alternative approaches to estimate long-range optical flow. 

\Paragraph{Direct estimation.}One of the naive methods is to directly estimate long-range flow with two distant reference images. Other than RAFT and GMA, we also compare the GMFlow~\cite{xu2022gmflow} which formulates the optical flow as a global matching problem to solve large motion. For fair comparisons, we also fine-tune the RAFT and GMA with training set of CVO, denoted as RAFT$^*$ and GMA$^*$, respectively.

\Paragraph{Pixel tracking.}Another intuitive way is to use pixel tracking method to iteratively estimate the per-pixel long-range displacement. We use the SOTA pixel tracking method PIPs~\cite{harley2022particle} to achieve this. Such process is time-consuming so we only test this method on CVO testing set.

\Paragraph{Warm start.} Zachary~\etal~\cite{raft2020} propose to estimate optical flow with warm start. This method can also be applied in flow accumulation, that is, we use the pre-obtained $\mathbf{F}_{1, t}$ as an initialized flow input to estimate $\mathbf{F}_{1, t+1}$. This procedure is essentially an implicit forward accumulation process, thus we include it into comparisons.

\subsection{Comparisons with Existing Methods}
\label{subsec:experiment_existing_method}

We compare the existing methods in terms of the average End-Point-Error (EPE) applied to all pixels (ALL) and occlusion regions (OCC). In Table~\ref{tab:main},
we compare our AccFlow with previous methods on two benchmarks, and our AccFlow outperforms all the previous methods by a large margin especially for occluded regions. 
Specifically, we notice that it is challenge for direct methods (the 1,5,9,13,18\textit{-th} rows in Table~\ref{tab:main}) to produce long-range optical flow due to the extreme large motion and occlusion problems. 
For forward accumulation, the explicit methods (\ie,~\cite{LimAG05} and~\cite{janai2017slow}) fail to handle the constantly increased occlusion which result in inferior performance. 
PIPs can accurately estimate sparse motion but suffers from the lack of spatial coherence information for dense flow estimation. 
Moreover, the implicit forward accumulation method (\ie, warm start) is not specially designed for this task and fall short in tackling occlusion problem, but still brings certain performance gain compared with direct methods.
Compared to all these methods, the AccFlow framework can decrease the average EPE error by large margin, which justifies the effectiveness of our framework for occlusion correction and non-occlusion correspondence enhancement. 
 
\begin{figure}[tbp]
   \centering\includegraphics[width=0.9\linewidth]{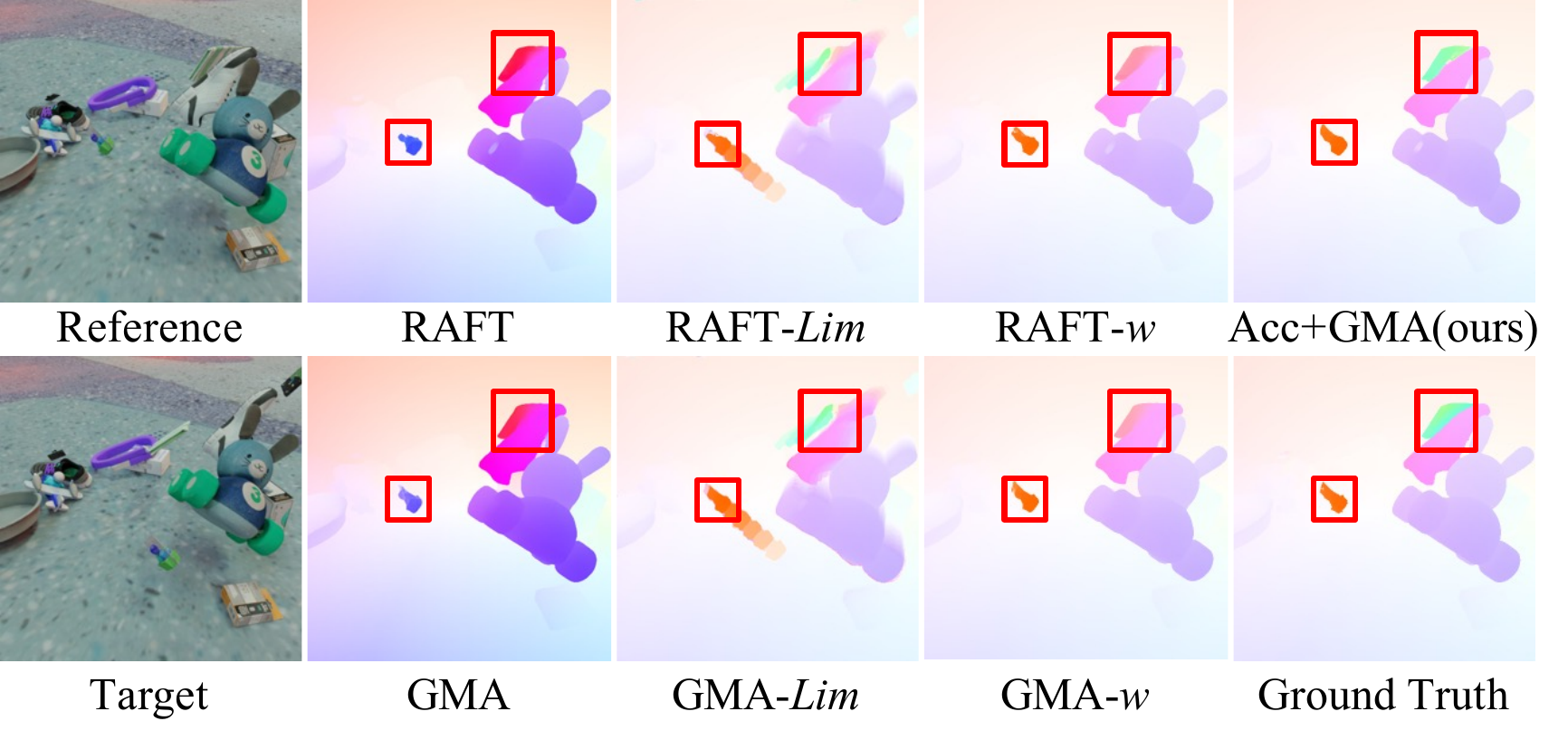}
   \caption{Visual quality comparisons on CVO dataset. Two small objects with large motions are emphasized with red boxes. More results can be found in the supplementary.}
   \label{fig:visual_cvo}
\end{figure}

Moreover, the qualitative comparisons are shown in Figure~\ref{fig:visual_cvo}, where two small objects with large motion are annotated in red boxes. It can be seen that our AccFlow can produce accurate optical flows while the compared methods suffer from significant errors especially for occluded area.


\begin{table}[t]
	\centering
	\resizebox*{ 0.9\linewidth}{!}{
		\begin{tabular}
			{   >{\centering\arraybackslash}p{0.8cm} 
				>{\centering\arraybackslash}p{0.8cm}| 
                >{\centering\arraybackslash}p{0.8cm}| 
				>{\centering\arraybackslash}p{0.7cm} 
                >{\centering\arraybackslash}p{0.7cm} 
				>{\centering\arraybackslash}p{0.7cm}| 
    			>{\centering\arraybackslash}p{0.7cm} 
                >{\centering\arraybackslash}p{0.7cm} 
				>{\centering\arraybackslash}p{0.7cm} 
			}
			\hline
			\multicolumn{2}{c|}{Acc+RAFT}  &\multirow{2}{*}{AB} & \multicolumn{3}{c|}{HS-Sintel}  & \multicolumn{3}{c}{CVO (Final)}   
			\\
			F.& B.& & ALL & NOC &  OCC & ALL & NOC&  OCC       \\
			\hline

            \checkmark& &           & 2.238 & 5.758  & 5.758  & 3.328 & 1.914  & 7.716  
			\\
            & \checkmark&           & 1.740 &1.303 & 4.711  & 2.709 &1.252 & 7.299  
			\\
            \checkmark& & \checkmark          & 1.716  &0.936 & 5.895  & 3.229 &0.873 & 8.823  
			\\
            & \checkmark& \checkmark& \textcolor{red}{1.383}  &\textcolor{red}{0.930} & \textcolor{red}{4.546}  & \textcolor{red}{2.707} &\textcolor{red}{1.249} & \textcolor{red}{7.295}  
			\\
			\hline



			\hline
	\end{tabular}}
    \caption{
		Ablation study of AccFlow framework (reported in EPE~$\downarrow$).  `F.' denotes a modified AccPlus that accumulates the local flows in forward manner, `B.' is the proposed AccPlus with backward accumulation, and `AB' denotes the adaptive blending module.
		}
    \vspace{-0.4cm}
	\label{tab:ablation}
\end{table}

\subsection{Ablation Study}
\label{subsec:ablation}

\Paragraph{Backward VS. Forward accumulation:} 
In Section~\ref{subsec:backward}, we demonstrate that the backward accumulation is less susceptible to occlusion effect than the forward one. In order to fairly compare the two methods, we design a modified AccPlus module which implements the forward accumulation in Table~\ref{tab:ablation} (denoted as `F.'). It is worth noting that the modification only change the inputs of network and no additional computational complexity is introduced. Detailed structure of the forward version of AccPlus is provided in appendix. In Table~\ref{tab:ablation}, we compare the backward accumulation with the forward one in terms of EPE under the same experimental settings. We can find that the backward version can deal with the occluded area more effectively than the forward version by large margin. This is because the backward accumulation has stable and minimum occlusion proportion at each step of iterations.

\Paragraph{Adaptive blending module:} 
In Section~\ref{subsec:accflow}, we design the AccFlow framework not only to address occlusion problem but also suppress the accumulation error. Specifically, the adaptive blending module takes a directly estimated long-range flow as prior to rectify the cumulated flow. To evaluate this, we train networks w/.~and w/o.~the adaptive blending module (denoted as `AB') in Table~\ref{tab:ablation}. 
The EPE is reduced by large margin especially for non-occluded area (NOC), which demonstrates the necessity of adaptive blending module for mitigating accumulated error.

\begin{figure}
   \centering\includegraphics[width=0.8\linewidth]{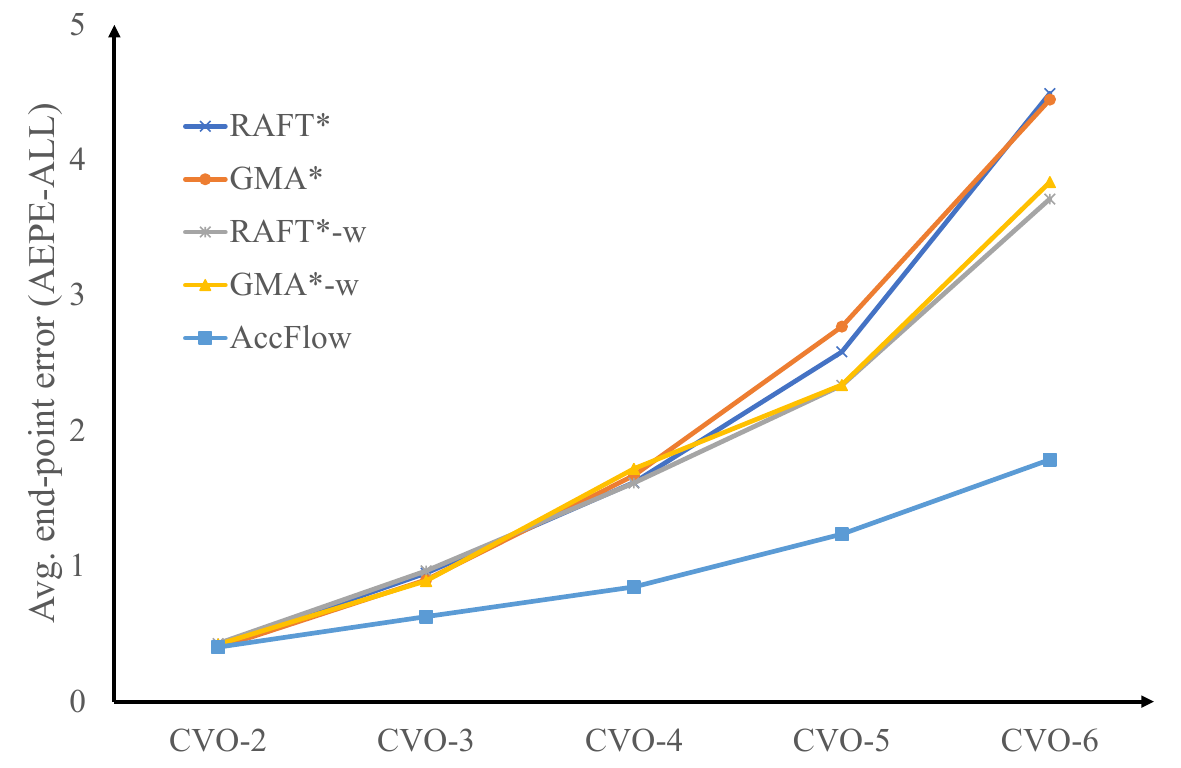}
   \caption{Average EPE~$\downarrow$ (ALL) of long-range flows from the compared methods in different estimation ranges.}
   \vspace{-0.4cm}
   \label{abl:cvo_step}
\end{figure}

\Paragraph{Accumulation for different frame ranges} 
In Figure~\ref{abl:cvo_step}, we show the results of long-range optical flow estimation in different estimation ranges. When the range increases, the EPE of the flows from our proposed AccFlow (Acc+GMA$^*$) increases slower than that from direct estimation and the warm start methods. This observation shows the robustness of our proposed framework in different estimation ranges. 

\section{Conclusion}
\label{sec:conclusion}
We propose the backward accumulation strategy for improved long-range optical flow estimation, surpassing prior methods. AccFlow employs feature domain backward accumulation and DNN-based error correction. Experimental results effectively address occlusion and accumulation errors. Ablation studies confirm superiority and adaptive blending's necessity. AccFlow notably reduces EPE on several benchmarks. In conclusion, AccFlow offers a simple, potent solution for flow accumulation, with scalability.

\section{Acknowledgment}
\label{sec:acknowledgment}
The work was supported in part by the Shanghai Pujiang Program under Grant 22PJ1406800, in part by the GuangDong Basic and Applied Basic Research Foundation under Project (No.2023A1515010644), and in part by Sichuan Provincial Key Laboratory of Intelligent Terminals under Grant SCITLAB-20016.

\clearpage

{\small
\bibliographystyle{ieee_fullname}
\bibliography{egbib}
}

\end{document}